\pdfoutput=1

\documentclass{article}

\usepackage{microtype}
\usepackage{graphicx}
\graphicspath{{figures/}}
\usepackage{subcaption}
\usepackage{booktabs}
\usepackage{hyperref}

\usepackage[preprint]{icml2026}

\usepackage{amsmath}
\usepackage{amssymb}
\usepackage{mathtools}

\usepackage[capitalize,noabbrev]{cleveref}

\icmltitlerunning{Hyperbolic Latent Geometry for Tree-Structured Prototype Networks}

\begin{document}

\twocolumn[
\icmltitle{Hyperbolic Latent Geometry for Tree-Structured\\
  Prototype Networks: A Local-vs-Global Trade-off}

\icmlsetsymbol{equal}{*}

\begin{icmlauthorlist}
\icmlauthor{Peter Flo}{harvard}
\icmlauthor{Luca Grossmann}{harvard}
\end{icmlauthorlist}

\icmlaffiliation{harvard}{Harvard University, Cambridge, MA, USA}

\icmlcorrespondingauthor{Peter Flo}{pgrindehollevik@g.harvard.edu}

\icmlkeywords{hyperbolic geometry, structured regularization, hierarchical classification, prototype networks, Poincar\'e ball}

\vskip 0.3in
]

\printAffiliationsAndNotice{}

\begin{abstract}
We study a tree-structured regularizer over class-prototype layouts
in a hierarchical-classification model and ask whether the choice of
latent manifold for the prototypes (Euclidean $\mathbb{R}^d$ vs.\
the Poincar\'e ball $\mathbb{B}^d_c$) affects how well that
regularizer can be satisfied without distorting the data
likelihood. The two manifolds differ only in
their volume growth: hyperbolic space grows exponentially with radius
and embeds trees with provably lower distortion than $\mathbb{R}^d$ of
matched dimension, so the structured regularizer should be cheaper to
satisfy on $\mathbb{B}^d_c$. Across 150 seed-replicated regularized
maximum-likelihood fits spanning embedding dimension, curvature, and
regularizer strength on WikiArt (27 styles, 81{,}446 paintings,
frozen CLIP ViT-B/16 features), we find a single robust effect:
Poincar\'e prototypes preserve the topology of the nearest-neighbor
graph in latent space substantially better than matched Euclidean
prototypes (sibling recall@5 $+8.7$\,pp,
cousin recall $+15.2$\,pp; paired-$t$ $p<10^{-4}$, sign agreement
$0.94$), and the gap holds across three reference-tree definitions
(hand-built lineage, CLIP-derived, and DINOv2-derived). On
classification, Euclidean prototypes are tied with logistic regression on raw
encoder features, indicating no detectable
contribution from the latent geometry; only the hyperbolic fit
improves on a $k$-NN encoder baseline for local retrieval. Global
tree-fidelity comparisons are unstable across reference trees and we
do not claim a winner. The results give an empirical separation, on a
real hierarchical-classification problem, between two natural latent
geometries for a class-structured regularizer.
\end{abstract}

\section{Introduction}
\label{sec:intro}

Many real-world classification problems carry an externally specified
hierarchy over class labels: artistic-style movements branch into one
another \citep{tan2019wikiart}, biological taxa nest into clades, and
ontologies of products, diseases, and documents are organised as
trees. A natural way to fold that hierarchy into a probabilistic
classification model is to add a tree-structured regularizer to the
layout of class-conditional parameters. We study the simplest version
of this: a prototype classifier whose class-conditional likelihood is
a softmax over distance-to-prototype, with a regularizer on the
matrix of pairwise prototype distances pulling it toward the
tree-distance matrix. The fitted classifier is then a balance between
a class-likelihood term (paintings of style $k$ should be close to
prototype $\mathbf{p}_k$) and a tree-shape term (prototypes
themselves should respect the hierarchy).

The choice of latent manifold for the prototypes is the question this
paper investigates. Forcing a tree's exponentially-growing leaf count
into Euclidean $\mathbb{R}^d$, whose ball volume grows only
polynomially in radius, induces unavoidable distortion of pairwise
distances; this distortion has been characterised analytically and
shown to drop sharply on the Poincar\'e ball $\mathbb{B}^d_c$, whose
own exponential volume growth matches a tree's
\citep{nickel2017poincare,sala2018representation}. In our model, the
tree-structured regularizer is a function of pairwise prototype
distances and should therefore be cheaper to satisfy on the
hyperbolic manifold than on the Euclidean one without distorting the
class-conditional likelihood. Whether that geometric advantage
realises in practice on a real hierarchical-classification problem is
what we test.

We instantiate the question on WikiArt-Refined \citep{tan2019wikiart},
a 27-style 81{,}446-painting corpus where the class hierarchy
(Renaissance branches into Baroque, Baroque into Rococo, Rococo into
Romanticism, on into Impressionism and Cubism) is part of the domain
and was hand-built from standard art-history references. Frozen
CLIP ViT-B/16 features \citep{radford2021clip} feed an MLP head and a
prototype classifier; the manifold of the classifier output is the
only design choice that varies between Euclidean and hyperbolic runs.
The same tree-structured regularizer (a normalised squared-error
pull on the prototype distance matrix) applies in both geometries.

We sweep regularized maximum-likelihood fits across 150
seed-replicated configurations spanning embedding dimension
$d \in \{2, 4, 8, 16, 32, 64\}$, curvature $c \in \{0.1, 0.3, 1, 3\}$,
regularizer strength $\lambda \in \{0, 0.1, 0.3, 1, 3\}$, and three
reference-tree variants (default lineage, chronological era grouping,
flat null). Three empirical findings emerge.

\paragraph{Local hierarchy.} Hyperbolic prototypes recover local tree
structure substantially better than matched Euclidean ones. Sibling
recall@5 at $d{=}8$ is $0.195$ versus $0.149$ on the default tree;
aggregated over 18 paired seed configurations the mean gap is
$+8.7$\,pp on sibling recall and $+15.2$\,pp on cousin recall, with
sign agreement $0.94$ and paired-$t$ $p<10^{-4}$. This effect is
preserved when sibling sets are redefined from data-driven empirical
trees built either from CLIP or DINOv2 \citep{oquab2024dinov2}
features, and across the regularizer-strength sweep.

\paragraph{Calibration.} A logistic-regression baseline on raw CLIP
features achieves $64.1\%$ top-1, statistically tied with the
Euclidean fit's $64.3\%$. A $k$-NN baseline on the same features
achieves sibling recall@5 of $0.160$, statistically tied with the
Euclidean fit's $0.149$ but $3.5$\,pp below the hyperbolic fit's
$0.195$. Read against these baselines, the hyperbolic fit is the only
one of our trained models that improves the encoder on local
hierarchy; the Euclidean fit matches the encoder on classification
but adds no detectable structural value over it.

\paragraph{Global tree fidelity.} Mean tree distortion against the
default tree favours Euclidean significantly, but
prototype-tree Spearman is not significantly different from zero
across the full sweep ($p=0.68$), and the small effect that does
appear flips sign between empirical reference trees built from CLIP
and from DINOv2 feature centroids. We therefore do not claim a
geometry winner on global tree fidelity.

The clean claim our experiments support is narrower than ``geometry
matters globally'': hyperbolic latent structure adds local-retrieval
value over the encoder while Euclidean latent structure does not, and
that local advantage is robust to which of three reference trees
defines ``local.''

\paragraph{Related work.} Hyperbolic prototype embeddings are an
established tool for tree-shaped data
\citep{nickel2017poincare,sala2018representation,ganea2018hyperbolic}.
\citet{khrulkov2020hyperbolic} report a similar asymmetry on
few-shot benchmarks (helps neighborhood structure, leaves top-1
unchanged); we confirm that pattern on a hand-built taxonomy with a
tree-aware regularizer, and quantify it against linear and $k$-NN
calibration baselines that previous reports of the same asymmetry do
not include. Tree-distance-matrix penalties over learnable points are
used implicitly in hyperbolic hierarchical clustering
\citep{chami2020trees}; here we use one as an explicit, tunable
regularizer.

\section{Data and reference trees}
\label{sec:data}

We use WikiArt-Refined \citep{tan2019wikiart}, $\sim$81{,}446
paintings labelled with one of 27 styles, with the supplied 70/30
train/val split. The corpus is heavily class-imbalanced
($133.3\times$ ratio between Impressionism with 13{,}060 examples and
Analytical Cubism with 77; full distribution in
Appendix~\ref{app:data}). Two design choices follow: the prototype
classifier has no per-class bias (only geometric distance enters the
logit), and we report balanced accuracy alongside top-1.

\paragraph{Reference trees.} WikiArt ships no hierarchy and
art-history offers several credible taxonomies, so we use three. The
\emph{default} tree (\cref{app:tree}) follows standard
art-historical lineage. The \emph{chronological} tree groups styles
into six era buckets (Renaissance, Baroque--Rococo, 19th century,
early 20th century, modern post-war, non-Western). The \emph{flat}
tree makes every style a direct child of the root and serves as a
deliberate null whose pairwise distances are constant off-diagonal.
For a tree $T$ with leaves $u, v$, tree distance is the unweighted
shortest-path edge count
$d_T(u, v) = \mathrm{depth}(u) + \mathrm{depth}(v) - 2\,\mathrm{depth}(\mathrm{LCA}(u, v))$.
We additionally build two \emph{empirical} reference trees by
agglomerative clustering of class-mean encoder features
(\cref{ssec:local}): one from CLIP and one from DINOv2, used only
for evaluation.

\section{Model and estimation}
\label{sec:methods}

\paragraph{Likelihood model.} Let $\phi$ be a frozen CLIP ViT-B/16
encoder \citep{radford2021clip} and $g_\theta : \mathbb{R}^{512} \to
\mathcal{M}$ a two-layer MLP with GELU and dropout, where the latent
manifold $\mathcal{M}$ is either Euclidean $\mathbb{R}^d$ or a
Poincar\'e ball $\mathbb{B}^d_c$ of curvature $c > 0$. For an image
$x$ of style $y \in \{1, \ldots, K\}$, set $z = g_\theta(\phi(x))$,
and let $\{\mathbf{p}_k\}_{k=1}^K \subset \mathcal{M}$ be learnable
class prototypes. The class-conditional likelihood is a softmax over
distance to prototype on $\mathcal{M}$:
\[
  p(y = k \mid z;\, \{\mathbf{p}_k\})
    \;\propto\; \exp\!\bigl(-d_\mathcal{M}(z, \mathbf{p}_k)\bigr).
\]
The encoder feeds the head identically in both geometries; switching
geometries changes exactly two things: the head's final transform
(identity for $\mathbb{R}^d$; $\exp_0^c(u) = \tanh(\sqrt{c}\,\|u\|)\,
u/(\sqrt{c}\,\|u\|)$ for the ball) and the metric used by the
classifier:
\[
  d^c(x, y) = \tfrac{1}{\sqrt{c}}\operatorname{arcosh}\!\Bigl(
    1 + \tfrac{2c\,\|x-y\|^2}{(1-c\|x\|^2)(1-c\|y\|^2)}
  \Bigr).
\]
Backbone, MLP width, dropout, batch size, optimiser, and schedule
are shared.

\paragraph{Tree-structured regularizer.} We add a tree-structured
regularizer on prototype layouts. Let
$D_{ij} = d_\mathcal{M}(\mathbf{p}_i, \mathbf{p}_j)$ be the matrix of
pairwise prototype distances on $\mathcal{M}$ and
$T_{ij} = d_T(i, j)$ the tree distance matrix on the reference tree
$T$. The regularizer penalises deviation between the \emph{shapes} of
these two distance matrices,
\[
  \mathcal{R}(\{\mathbf{p}_k\}) \;=\;
    \frac{\lambda}{|\mathcal{P}|}
    \sum_{(i,j) \in \mathcal{P}}
    \Bigl(\tfrac{D_{ij}}{\bar{D}} - \tfrac{T_{ij}}{\bar{T}}\Bigr)^{\!2},
\]
where $\mathcal{P}$ is the set of $\binom{K}{2}$ off-diagonal pairs,
$\bar{D}, \bar{T}$ are the means of $D, T$, and $\lambda$ controls
regularizer strength. Mean-normalising each pair before differencing
makes the regularizer invariant to the absolute scale of distances on
$\mathcal{M}$: the geometry is free to pick whatever scale the
likelihood prefers, while the regularizer constrains only the
relative structure of the prototype distance matrix. $\lambda = 0$
recovers the cross-entropy baseline used in prior hyperbolic-image
work \citep{khrulkov2020hyperbolic}. We treat $\mathcal{R}$ as a
penalty rather than a Bayesian log-prior: $\mathcal{R}$ is not the
log of a normalised density on $\mathcal{M}^K$ (it is mean-rescaled
on each evaluation), so it has no proper-prior interpretation, and we
report regularized maximum-likelihood estimates rather than MAP
estimates of a posterior.

\paragraph{Estimation.} We minimise the regularized negative
log-likelihood
$\mathcal{L}(\theta, \{\mathbf{p}_k\}) = \mathcal{L}_{CE}(\theta, \{\mathbf{p}_k\}) + \mathcal{R}(\{\mathbf{p}_k\})$
by stochastic gradient descent. Adam optimises the MLP head;
Riemannian Adam from \texttt{geoopt} \citep{kochurov2020geoopt}
optimises the hyperbolic prototypes, projecting each update back onto
the manifold. The regularizer is computed once per gradient step over
the $\binom{27}{2} = 351$ off-diagonal style pairs, negligible cost
next to the per-batch likelihood term. The framing makes explicit
that the regularizer is a structured penalty on a geometric quantity
(the prototype distance matrix) and that the choice of latent
manifold controls how cheaply that penalty can be driven to zero.

\paragraph{Distortion bound.} The regularizer is a
function of pairwise prototype distances, so in any latent geometry
where the tree distance matrix $T$ admits a low-distortion isometric
embedding, the regularizer cost at the optimum is small. Sala
et~al.\ give distortion bounds $\sim 1/d$ for $\mathbb{B}^d_c$ versus
a constant floor for $\mathbb{R}^d$ at fixed $d$
\citep{sala2018representation}. The empirical question is whether
that asymptotic statement is visible on a real dataset at modest
$d$, against a real classification likelihood, with a finite-sample
fit rather than an isometric embedding objective.

\paragraph{Training details.} 150 runs total; batch size 4096; Adam
$\eta = 10^{-3}$ for the head and $10^{-2}$ for the prototypes;
weight decay $10^{-4}$; dropout $0.1$; gradient clipping at norm
$1.0$; 30 epochs. CLIP features are pre-cached as float16 tensors,
keeping each run under 10 seconds on a single Apple M-series GPU and
the full sweep under half an hour. Every headline configuration is
replicated across three seeds; reported error bars are seed standard
deviation. Significance comparisons across seeds use paired-$t$ tests
on the 18 (dim, seed) pairs of the dimension sweep, with hyperbolic
configurations selected at the best curvature per pair.

\paragraph{Evaluation.} A useful style embedding can be useful along
three orthogonal axes, and we report metrics for each.
\emph{Classification} is top-1, top-5, and balanced accuracy.
\emph{Global tree fidelity} is the Spearman correlation between the
learned prototype distance matrix and the tree distance matrix, plus
mean and worst-case multiplicative distortion. \emph{Local tree
preservation} is sibling and cousin recall@$k$ among the $k$ nearest
neighbours of each validation embedding, for $k \in \{5, 10\}$.
Sibling/cousin sets are derived from a reference tree; we evaluate
against the default, the CLIP-empirical, and the DINOv2-empirical
tree to test sensitivity (\cref{ssec:local,ssec:robust}).

\section{Results}
\label{sec:results}

Three findings emerge from the 150-run sweep, summarised here and
developed below.
(i)~Hyperbolic prototypes recover local tree structure substantially
better than matched Euclidean ones, robustly across reference-tree
construction (\cref{ssec:local}).
(ii)~Calibrated against linear and $k$-NN baselines on raw CLIP
features, the hyperbolic fit is the only trained model that
adds structural value over the encoder; the Euclidean fit
matches the encoder on classification but adds nothing on retrieval
(\cref{ssec:calib}).
(iii)~The global tree-fidelity comparison is metric- and
reference-tree-dependent and we do not claim a winner there
(\cref{ssec:global}).

\begin{figure*}[t]
  \centering
  \includegraphics[width=0.85\linewidth]{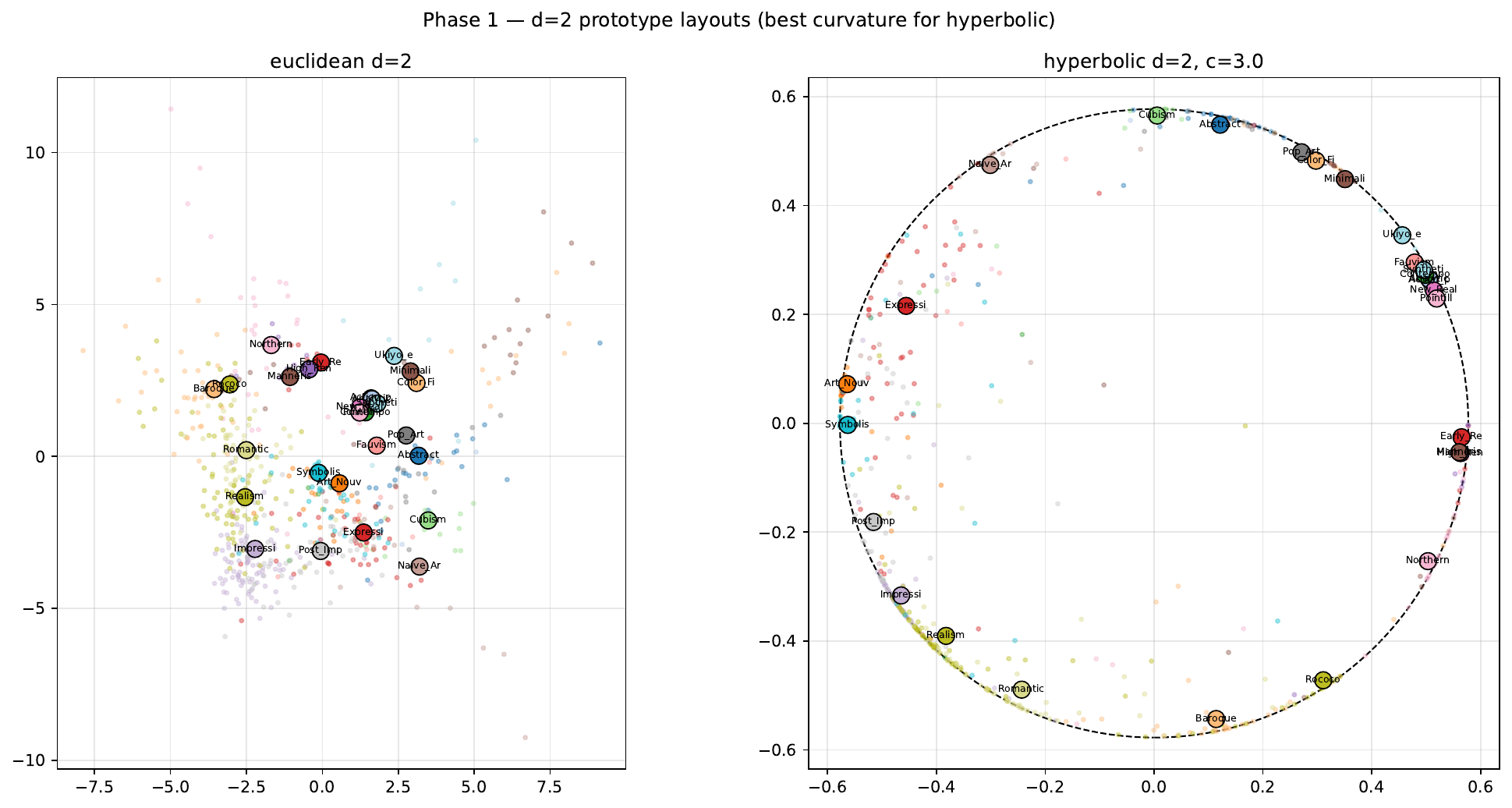}
  \caption{Prototype layouts at $d{=}2$, best curvature per geometry.
  Each large dot is a learned style prototype; small dots are a
  600-image sample of validation embeddings. The dashed circle on
  the right marks the Poincar\'e ball boundary at radius
  $1/\sqrt{c}$. Hyperbolic prototypes settle into a near-boundary
  ring (mean radius $0.561$, std $0.013$ across all 27 styles, against
  a boundary at $0.577$). This is the exponential-volume regime
  where the manifold has the capacity for trees \citep{nickel2017poincare,
  sala2018representation}. Euclidean prototypes scatter diffusely
  with no analogous structural pressure (mean radius $2.96$, std
  $0.995$). The geometry behaves as theory predicts; the rest of
  the paper measures whether that geometric difference shows up in
  downstream metrics.}
  \label{fig:disk}
\end{figure*}

\begin{table*}[t]
  \centering
  \small
  \begin{tabular}{lcccc}
\toprule
 & euclidean & hyperbolic & logistic-on-CLIP & kNN-5-on-CLIP \\
metric &  &  &  &  \\
\midrule
Top-1 & 0.659 ± 0.003 & 0.606 ± 0.002 & 0.641 & 0.635 \\
Top-5 & 0.959 ± 0.001 & 0.935 ± 0.001 & — & — \\
Balanced acc. & 0.554 ± 0.002 & 0.463 ± 0.004 & 0.611 & 0.588 \\
Class-center / tree Spearman & 0.350 ± 0.027 & 0.242 ± 0.002 & — & — \\
Avg. tree distortion & 1.742 ± 0.009 & 1.861 ± 0.003 & — & — \\
Worst tree distortion & 4.733 ± 0.253 & 7.147 ± 0.076 & — & — \\
Dendrogram F1 & 0.034 ± 0.030 & 0.000 ± 0.000 & — & — \\
Sibling recall@5 & 0.136 ± 0.003 & 0.188 ± 0.004 & — & 0.160 \\
Cousin recall@5 & 0.249 ± 0.003 & 0.296 ± 0.005 & — & 0.277 \\
\bottomrule
\end{tabular}

  \caption{Best Euclidean vs.\ best hyperbolic configuration across
  the dimension and regularizer-strength sweeps, selected per geometry by
  mean top-1 across seeds. Mean $\pm$ seed standard deviation. Two
  rightmost columns: logistic regression and $k$-NN-5 on raw CLIP
  features (no learned head, no prototype objective). Euclidean
  leads classification and global-tree metrics over hyperbolic
  prototypes; hyperbolic prototypes lead sibling and cousin recall
  by $4$--$5$\,pp. Read against the calibration baselines, the
  Euclidean fit is tied with logistic on top-1 and
  with $k$-NN on sibling recall; only the hyperbolic fit improves
  on $k$-NN for sibling and cousin recall.}
  \label{tab:headline}
\end{table*}

\subsection{Headline comparison}

Selecting the best configuration of each geometry by mean top-1
yields \cref{tab:headline}: the two geometries split along a
local-vs-global axis. Euclidean leads classification (top-1
$+5.3$\,pp, top-5 $+2.4$\,pp) and global-tree Spearman against the
default tree ($+0.108$); hyperbolic leads sibling recall@5
($+0.052$) and cousin recall@5 ($+0.047$). The seed bands do not
overlap on any of the five metrics. The remaining subsections show
that this split is structural rather than the artifact of any single
configuration.

\subsection{Local tree structure}
\label{ssec:local}

\begin{figure*}[t]
  \centering
  \includegraphics[width=0.95\linewidth]{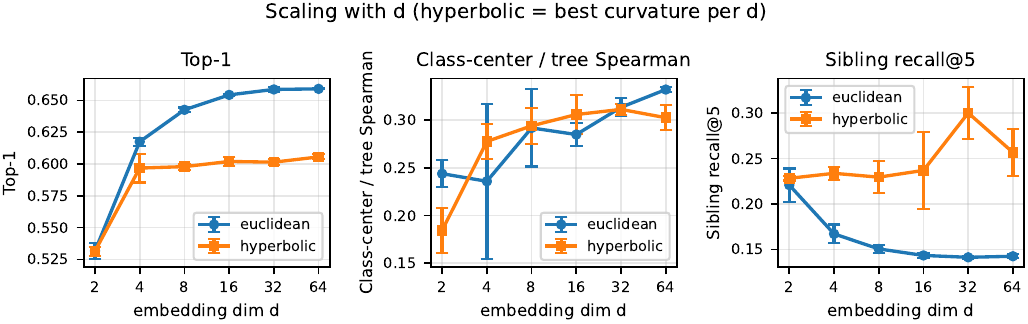}
  \caption{Metrics versus embedding dimension $d$ (90 runs; three
  seeds per config; hyperbolic at the best curvature per $d$ per
  seed). Top-1 saturates near $d{=}16$ for both geometries with
  Euclidean ahead by $4$--$6$\,pp. Sibling recall@5
  decreases with $d$ for Euclidean (from $0.221$ at $d{=}2$ to
  $0.142$ at $d{=}64$) but stays nearly flat for hyperbolic,
  opening a $5$\,pp gap by $d{=}64$.}
  \label{fig:phase1}
\end{figure*}

\begin{figure*}[t]
  \centering
  \includegraphics[width=0.95\linewidth]{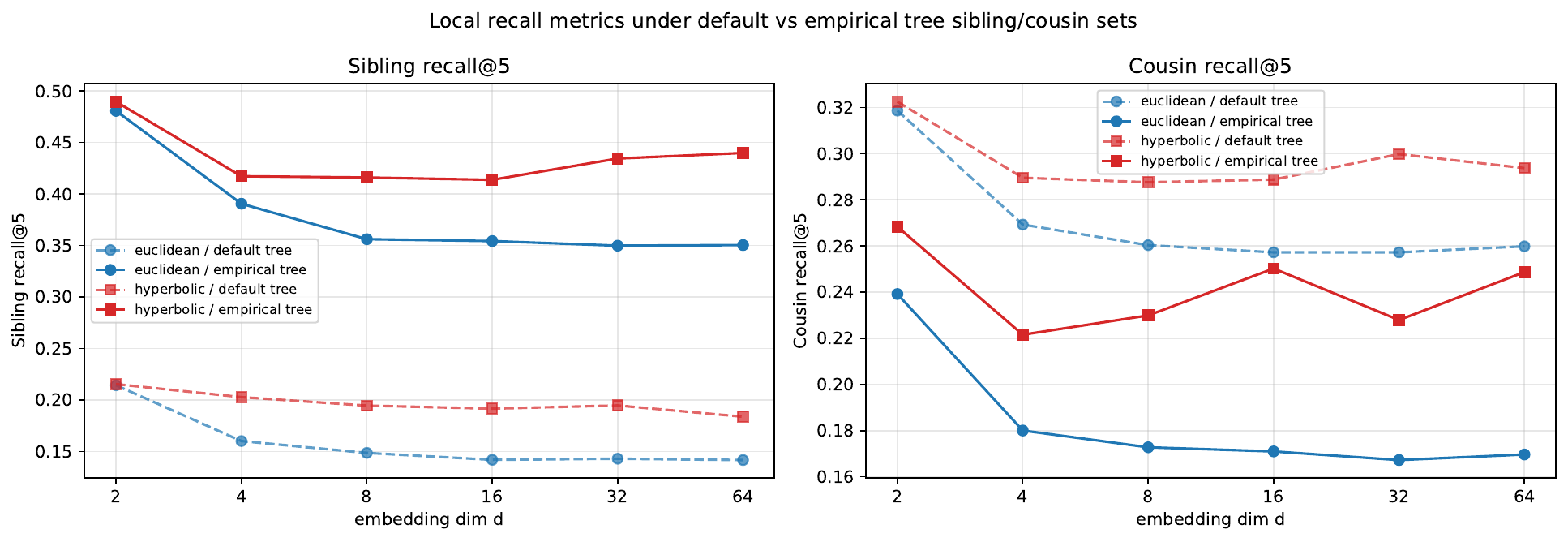}
  \caption{Sibling and cousin recall@5 with relations defined by the
  default tree (dashed) and a CLIP-derived empirical tree (solid),
  per geometry per embedding dimension. The hyperbolic lead is
  preserved under the empirical-tree redefinition; on sibling recall
  the gap widens slightly. The DINOv2-derived empirical tree is in
  \cref{app:phase45}.}
  \label{fig:empirical_recall}
\end{figure*}

The local-recall difference is the strongest signal in the sweep.
On the default tree (\cref{fig:phase1}), hyperbolic sibling recall@5
stays near $0.19$ across $d \in \{2, 4, 8, 16, 32, 64\}$, while
Euclidean sibling recall decreases from $0.221$ at $d{=}2$ to
$0.142$ at $d{=}64$. This monotone divergence with embedding
capacity is consistent with the geometric prediction: in
$\mathbb{R}^d$ the distortion of the tree distance matrix grows with
the available capacity to spread classes for likelihood, while in
$\mathbb{B}^d_c$ the curvature absorbs that pressure. Aggregated over 18 paired seed
configurations (6 dimensions $\times$ 3 seeds, hyperbolic at the
best curvature per pair), the mean gap is $+8.7$\,pp on sibling
recall and $+15.2$\,pp on cousin recall, with sign agreement $0.94$
on each and paired-$t$ $p<10^{-4}$.

The sibling/cousin sets used by recall@5 are themselves a function of
which reference tree we use, so a sceptical reading would treat the
local advantage as an artifact of the hand-built tree. We test this
by rebuilding sibling and cousin sets from data-driven
\emph{empirical} trees: agglomerative average-linkage clustering on
class-mean encoder features, converted to integer-edge tree
distances. \cref{fig:empirical_recall} shows the comparison
against a CLIP-derived empirical tree; the hyperbolic lead is
preserved and slightly larger ($+6.0$\,pp at $d{=}8$ versus
$+4.6$\,pp on the default tree). Replacing the empirical tree's
construction encoder with DINOv2 \citep{oquab2024dinov2} replicates
the lead at every $d \geq 4$ (full numbers in \cref{app:phase45}).
The local advantage of hyperbolic prototypes is reference-tree
independent in the strict sense that it survives three distinct
reference-tree constructions, including two encoder-derived ones
that share no construction step with the hand-built tree.

\subsection{Calibration baselines}
\label{ssec:calib}

\begin{figure}[t]
  \centering
  \includegraphics[width=\linewidth]{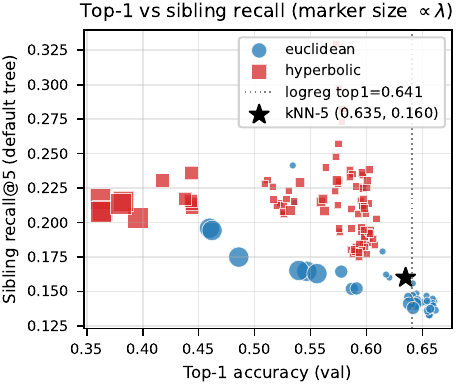}
  \caption{Top-1 accuracy versus sibling recall@5 for every sweep
  configuration. Marker size scales with the regularizer
  strength $\lambda$. Vertical dotted line: logistic-regression top-1 on raw
  CLIP features ($64.1\%$). Black star: $k$-NN-5 on raw CLIP
  features (sibling recall $0.160$). The Euclidean cluster sits at or
  below the encoder baseline on retrieval; only the hyperbolic
  cluster populates the upper region of the plot.}
  \label{fig:pareto}
\end{figure}

A reader cannot judge a $65.9\%$ top-1 number without external
calibration. Two reference classifiers train directly on raw frozen
CLIP features, with no learned head and no prototype objective:
logistic regression and $k$-NN-5. Logistic regression achieves
$64.1\%$ top-1 and $61.1\%$ balanced accuracy; $k$-NN achieves
$63.5\%$ top-1, $58.8\%$ balanced accuracy, and sibling recall@5 of
$0.160 / 0.366$ on default-tree / CLIP-empirical-tree relations.

Read against these baselines: the best Euclidean configuration
($65.9\%$ top-1) beats logistic regression
by $1.8$\,pp; on balanced accuracy it underperforms ($55.4\%$ vs.\
$61.1\%$). Best-top-1 Euclidean's sibling recall@5 is $0.136$,
below the $k$-NN baseline of $0.160$. The best-top-1
hyperbolic configuration ($60.6\%$ top-1) sits $2.9$\,pp below
$k$-NN on classification but $2.8$\,pp above it on sibling recall
($0.188$ vs.\ $0.160$). At $d{=}8$
specifically, hyperbolic sibling recall@5 reaches $0.195$ on the
default tree and $0.416$ on the empirical tree, against the $k$-NN
baseline's $0.160 / 0.366$. \cref{fig:pareto} plots the full
top-1-vs-sibling-recall scatter: the Euclidean cluster sits where
$k$-NN already lives; only hyperbolic configurations populate the
upper region. On this dataset and encoder, the hyperbolic fit
is the only trained model that adds local-structural value over the
encoder.

\subsection{Regularizer strength}
\label{ssec:reg}

\begin{figure*}[t]
  \centering
  \includegraphics[width=0.95\linewidth]{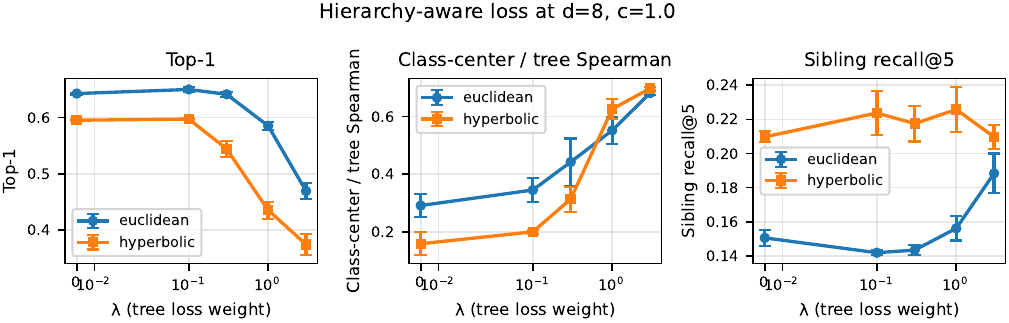}
  \caption{Top-1, prototype-tree Spearman, and sibling recall@5
  versus regularizer strength $\lambda$ at $d{=}8$, $c{=}1$ (three
  seeds per point). The regularizer drives tree-Spearman from
  $\sim 0.3$ to $\sim 0.7$ in both geometries, but classification
  collapses for $\lambda \geq 1$. The hyperbolic top-1 gap never
  closes; sibling recall is stable for hyperbolic across all
  $\lambda$, while for Euclidean it rises only at the largest
  tested $\lambda$, where classification has already collapsed.
  Mean tree distortion (omitted here for legibility) decreases
  uniformly with $\lambda$ in both geometries; full four-panel
  version in \cref{app:phase2-full}.}
  \label{fig:phase2}
\end{figure*}

The relative ordering of the two geometries is preserved across the
regularizer-strength sweep $\lambda \in \{0, 0.1, 0.3, 1, 3\}$ at
$d \in \{8, 16\}$, $c{=}1$ (\cref{fig:phase2}). At $\lambda{=}0.1$
the regularizer Pareto-improves both: Euclidean $d{=}8$ top-1 rises
$0.7$\,pp to $65.0\%$ while tree-Spearman rises from $0.292$ to
$0.345$; hyperbolic gains a fraction of a point on top-1 and
$2$\,pp on sibling recall. At $\lambda{=}3$ both geometries reach
tree-Spearman $\approx 0.7$ but classification collapses (Euclidean
$46.9\%$, hyperbolic $37.4\%$): the regularizer succeeds at the
metric it was constructed to optimise, at substantial likelihood
cost. The Euclidean top-1 lead and the hyperbolic sibling-recall
lead persist at every $\lambda$ tested.

\subsection{Training-tree and class-imbalance robustness}
\label{ssec:robust}

\begin{figure*}[t]
  \centering
  \includegraphics[width=0.95\linewidth]{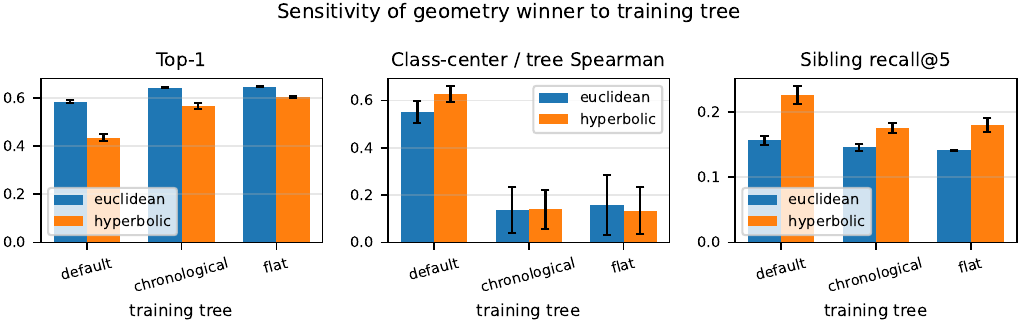}
  \caption{Each geometry trained with $\lambda{=}1$ at $d{=}8$
  against three different reference trees, then evaluated against
  the default tree. Bars are seed means; whiskers are seed standard
  deviation. Tree-Spearman against the default tree collapses for
  non-default training trees; the regularizer only helps the metric
  anchored to the training tree. The geometry winner does not change
  on any panel: Euclidean wins top-1 on every tree, hyperbolic wins
  sibling recall on every tree. Full four-panel version including
  mean tree distortion in \cref{app:phase3-full}.}
  \label{fig:phase3}
\end{figure*}

Re-training each geometry against the chronological and flat
reference trees (with $d{=}8$, $c{=}1$, $\lambda{=}1$, evaluated
against the default tree throughout) does not change the geometry
winner on any panel: Euclidean wins top-1 on every training tree,
hyperbolic wins sibling recall on every training tree
(\cref{fig:phase3}). Inverse-frequency class-weighted training
preserves the geometry gap on every metric we report (top-1 remains
$5$\,pp Euclidean, sibling recall remains $4$\,pp hyperbolic;
\cref{app:imbalance}). Substituting DINOv2 features into the
empirical-tree construction reverses the direction of the small
global-Spearman gap but preserves the hyperbolic local-recall lead
on every reference tree (\cref{app:phase45}).

\subsection{Global tree fidelity is unstable}
\label{ssec:global}

The third finding is partly a negative result. Aggregated paired-$t$
tests over the full dimension sweep give $p=0.68$ for class-center /
default-tree Spearman and a sign agreement of exactly $0.50$. The
small headline gap ($\rho_{\textrm{Eu}} - \rho_{\textrm{Hy}}\approx 0.108$
in \cref{tab:headline}) is therefore selecting a noisy outlier
through best-of-sweep. Mean tree distortion does favour Euclidean
significantly ($+0.040$, $p<10^{-4}$), so the two natural
global-fidelity metrics disagree about which geometry wins.
Replacing the hand-built reference with a CLIP-derived empirical
tree reverses the Spearman direction at every $d \geq 4$
(hyperbolic ahead by $+0.01$ to $+0.05$); replacing the empirical
tree's construction encoder with DINOv2 reverses it back (Euclidean
ahead by up to $+0.06$). Hand-built and encoder-derived trees
correlate only weakly ($\rho \approx 0.08, 0.12$ on pairwise
distances); CLIP-empirical and DINOv2-empirical correlate $0.50$.
Across all four reference-tree constructions we tried, the only
stable claim about global tree fidelity is that the comparison is
not stable. The local claim survives every variation.

\section{Discussion}
\label{sec:conclusion}

The clean finding is that, in our tree-regularized prototype model
on a real hierarchical-classification problem, the latent manifold
has a local effect on tree fidelity even when its global effect is
unstable. Hyperbolic prototype geometry is the only one of
our trained models that meaningfully improves on the frozen encoder
for local hierarchy preservation. Sibling recall@5 at $d{=}8$ is
$0.195$ for hyperbolic, $0.149$ for Euclidean, and $0.160$ for
$k$-NN-5 on raw CLIP features: the Euclidean fit is tied with the
encoder; the hyperbolic fit improves on it by
$3.5$\,pp. The advantage holds when sibling sets are redefined from
data-driven empirical trees built from CLIP or DINOv2 features, and
aggregates to $+8.7$\,pp on sibling and $+15.2$\,pp on cousin recall
over the full dimension sweep with paired-$t$ $p<10^{-4}$.

The classification gap is real but uninformative about geometry. A
logistic-regression head on raw CLIP features achieves $64.1\%$
top-1, which the Euclidean fit matches at $d{=}8$ ($64.3\%$) and
the hyperbolic fit underperforms by $4$--$6$\,pp. Neither fit is doing meaningful
classification work that a linear head on the same features cannot;
the difference between the two fits on this axis is best read as
``Euclidean prototype distances do not distort classification beyond
what the encoder already supports, while hyperbolic prototype
distances do.''

The global tree-fidelity comparison is the place where careful
framing matters most. Mean tree distortion favours Euclidean
significantly against the default tree; class-center / tree Spearman
is not significantly different from zero across the sweep
($p=0.68$); the small global-Spearman gap reverses sign between
empirical reference trees built from CLIP and from DINOv2 feature
centroids. ``Either geometry wins on global tree fidelity'' is not a
claim our experiments support. Sibling and cousin recall favour
hyperbolic on every reference tree we tried.

\paragraph{Limitations.} We cannot generalise beyond medium-scale,
Western-canon-heavy WikiArt with a frozen CLIP-ViT-B/16 encoder for
training; the local advantage we identify may not survive a
fine-tuned encoder or a hierarchy substantially deeper than the
$27$-leaf taxonomy used here. We cannot conclude that the hyperbolic
geometry does substantive work inside the head MLP, since only the
prototypes live on the manifold.

\paragraph{Extensions.} Three natural follow-ups in increasing cost.
A genuinely hyperbolic head along the lines of M\"obius layers
\citep{ganea2018hyperbolic} would let the manifold shape the
representation rather than only the decision boundary, and is the
cheapest test of whether the classification gap closes. The
empirical reference trees in this paper are built in Euclidean space
from encoder features. Building one in hyperbolic space, or by direct
tree-learning methods \citep{chami2020trees}, would close a
remaining circularity. Replication on a deeper hierarchy
(iNaturalist with WordNet, or a fine-grained subdivision of
WikiArt's largest classes) would test the limits of the local
advantage we observe.

\section*{Broader impact}
\label{sec:impact}

The system itself is small (a 27-way classifier on cached features),
but a few of its design choices have implications worth being explicit
about. The default reference tree is built from a Western,
lineage-based art-history canon. Every metric in this paper that
mentions ``the hierarchy'' is anchored to that taxonomy, which is
implicitly endorsed by anyone using the numbers. Our tree-variant
experiments are partial mitigation: they document how much each
conclusion depends on the choice of tree. A deployment in a museum or
classroom should treat the tree as configuration, not as a constant.
WikiArt is heavily biased toward European painting; the corpus has
27 styles but only one (Ukiyo-e) sits outside the European-and-American
canon, and East Asian ink-painting traditions that span centuries are
collapsed into that single label. A retrieval system trained on this
data will under-rank non-Western works for ambiguous queries.
Hyperbolic geometry, which tightens local neighborhoods, could
compound the effect by making those already-tight neighborhoods more
confident. We therefore do not recommend deploying
hyperbolic style embeddings for attribution or authentication tasks
without expert human review: $50$--$65\%$ top-1 accuracy over $27$
well-known styles is far below the threshold any serious provenance,
insurance, or legal decision should require.

\section*{Acknowledgements}

This work used Anthropic's Claude (Opus 4.7) as a research assistant.
Specifically, the model assisted with code scaffolding for experimental
infrastructure (sweep orchestration, plotting helpers, figure
generation), with brainstorming and concept clarification during
analysis, and with preliminary drafts of prose that the authors
substantially revised. All experimental design choices, the selection
of hypotheses to test, hyperparameter ranges, ground-truth tree
definitions, result interpretations, and the final text of this
manuscript reflect the authors' substantial original contribution.

\bibliographystyle{icml2026}
\bibliography{references}

@inproceedings{nickel2017poincare,
  title     = {Poincar\'{e} Embeddings for Learning Hierarchical Representations},
  author    = {Nickel, Maximilian and Kiela, Douwe},
  booktitle = {Advances in Neural Information Processing Systems},
  volume    = {30},
  year      = {2017}
}

@inproceedings{ganea2018hyperbolic,
  title     = {Hyperbolic Neural Networks},
  author    = {Ganea, Octavian and B\'{e}cigneul, Gary and Hofmann, Thomas},
  booktitle = {Advances in Neural Information Processing Systems},
  volume    = {31},
  year      = {2018}
}

@inproceedings{radford2021clip,
  title     = {Learning Transferable Visual Models From Natural Language Supervision},
  author    = {Radford, Alec and Kim, Jong Wook and Hallacy, Chris and Ramesh, Aditya and Goh, Gabriel and Agarwal, Sandhini and Sastry, Girish and Askell, Amanda and Mishkin, Pamela and Clark, Jack and others},
  booktitle = {Proceedings of the 38th International Conference on Machine Learning},
  pages     = {8748--8763},
  year      = {2021},
  publisher = {PMLR}
}

@inproceedings{tan2019wikiart,
  title     = {Improved {ArtGAN} for Conditional Synthesis of Natural Image and Artwork},
  author    = {Tan, Wei Ren and Chan, Chee Seng and Aguirre, Hernán and Tanaka, Kiyoshi},
  booktitle = {IEEE Transactions on Image Processing},
  volume    = {28},
  number    = {1},
  pages     = {394--409},
  year      = {2019},
  publisher = {IEEE}
}

@inproceedings{kochurov2020geoopt,
  title     = {Geoopt: {Riemannian} Optimization in {PyTorch}},
  author    = {Kochurov, Max and Karimov, Rasul and Kozlukov, Serge},
  booktitle = {ICML 2020 Workshop on Graph Representation Learning and Beyond (GRL+)},
  year      = {2020},
  note      = {arXiv:2005.02819}
}

@inproceedings{sala2018representation,
  title     = {Representation Tradeoffs for Hyperbolic Embeddings},
  author    = {Sala, Frederic and De Sa, Christopher and Gu, Albert and R\'{e}, Christopher},
  booktitle = {Proceedings of the 35th International Conference on Machine Learning},
  pages     = {4460--4469},
  year      = {2018},
  publisher = {PMLR}
}

@inproceedings{khrulkov2020hyperbolic,
  title     = {Hyperbolic Image Embeddings},
  author    = {Khrulkov, Valentin and Mirvakhabova, Leyla and Ustinova, Evgeniya and Oseledets, Ivan and Lempitsky, Victor},
  booktitle = {IEEE/CVF Conference on Computer Vision and Pattern Recognition (CVPR)},
  year      = {2020}
}

@inproceedings{chami2020trees,
  title     = {From Trees to Continuous Embeddings and Back: Hyperbolic Hierarchical Clustering},
  author    = {Chami, Ines and Gu, Albert and Chatziafratis, Vaggos and R\'{e}, Christopher},
  booktitle = {Advances in Neural Information Processing Systems},
  volume    = {33},
  year      = {2020}
}

@article{oquab2024dinov2,
  title   = {{DINOv2}: Learning Robust Visual Features without Supervision},
  author  = {Oquab, Maxime and Darcet, Timoth{\'e}e and Moutakanni, Th{\'e}o and Vo, Huy and Szafraniec, Marc and Khalidov, Vasil and Fernandez, Pierre and Haziza, Daniel and Massa, Francisco and El-Nouby, Alaaeldin and Assran, Mahmoud and Ballas, Nicolas and Galuba, Wojciech and Howes, Russell and Huang, Po-Yao and Li, Shang-Wen and Misra, Ishan and Rabbat, Michael and Sharma, Vasu and Synnaeve, Gabriel and Xu, Hu and J{\'e}gou, Herv{\'e} and Mairal, Julien and Labatut, Patrick and Joulin, Armand and Bojanowski, Piotr},
  journal = {Transactions on Machine Learning Research},
  year    = {2024},
  note    = {arXiv:2304.07193}
}

\newpage
\appendix
\onecolumn
\section{Default style hierarchy}
\label{app:tree}

The default reference tree, hand-built from standard art-history
references and used as the primary evaluation reference throughout
the paper.

\begin{verbatim}
Root
|-- Early_Renaissance
|   |-- Northern_Renaissance
|   `-- High_Renaissance
|       `-- Mannerism_Late_Renaissance
|           `-- Baroque
|               `-- Rococo
|                   `-- Romanticism
|                       |-- Realism
|                       |   |-- Contemporary_Realism
|                       |   `-- Impressionism
|                       |       `-- Post_Impressionism
|                       |           |-- Pointillism
|                       |           |-- Fauvism
|                       |           `-- Cubism
|                       |               |-- Analytical_Cubism
|                       |               `-- Synthetic_Cubism
|                       `-- Symbolism
|                           |-- Art_Nouveau
|                           `-- Expressionism
|                               `-- Abstract_Expressionism
|                                   |-- Action_painting
|                                   |-- Color_Field_Painting
|                                   `-- Minimalism
|-- Pop_Art
|   `-- New_Realism
|-- Ukiyo_e
`-- Naive_Art_Primitivism
\end{verbatim}

\section{Class distribution and tree-distance matrices}
\label{app:data}

\begin{figure}[h]
 \centering
 \includegraphics[width=0.7\linewidth]{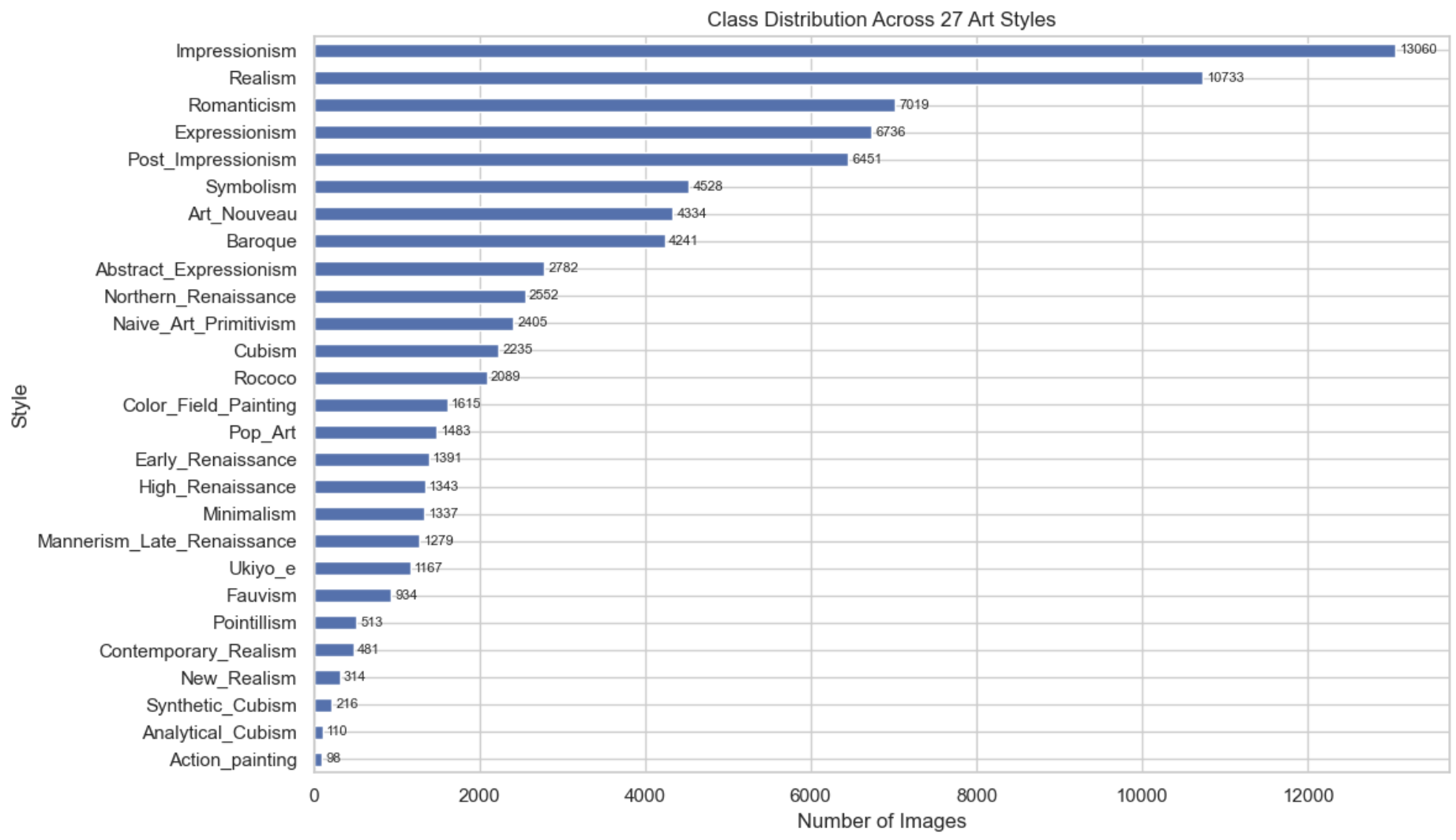}
 \caption{Per-style image counts on WikiArt-Refined. Impressionism
 dominates with $13{,}060$ examples; Analytical Cubism, Action
 Painting, and Synthetic Cubism each have fewer than $250$. The
 imbalance ratio between extrema is $133.3\times$.}
 \label{fig:class-distribution}
\end{figure}

\begin{figure}[h]
 \centering
 \includegraphics[width=\linewidth]{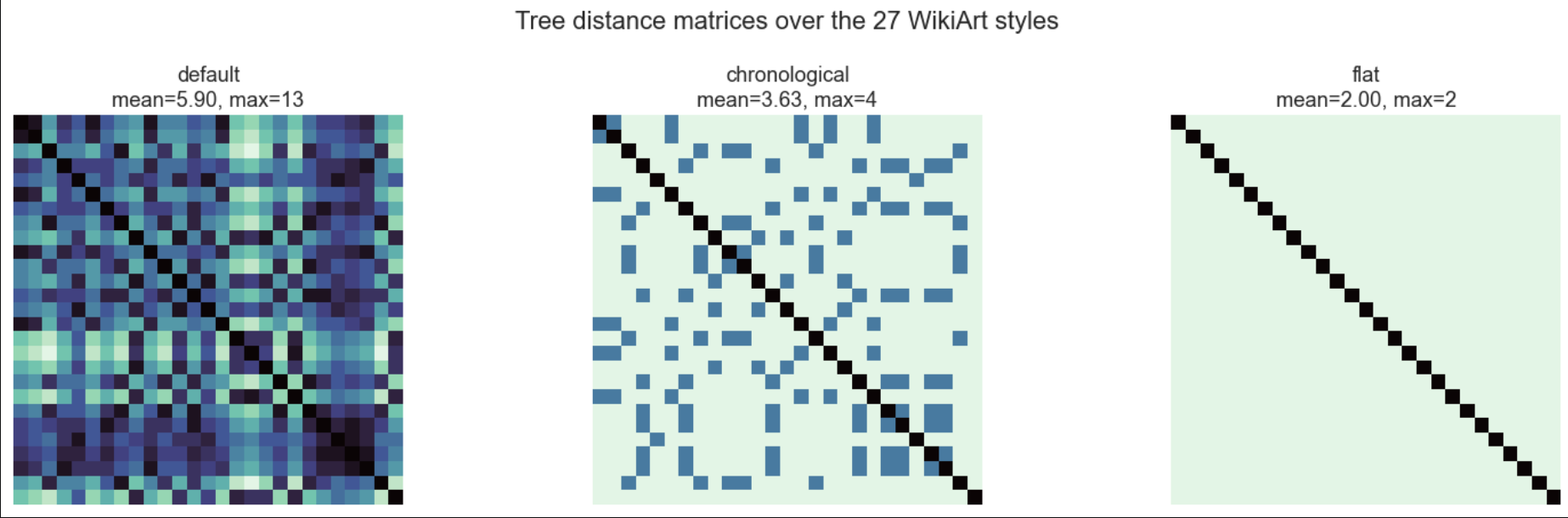}
 \caption{Tree distance matrices for the three reference hierarchies
 used in this paper: default lineage (left), chronological era
 grouping (centre), flat null (right). Rows and columns are styles in
 the same order across panels.}
 \label{fig:tree-distance-matrices}
\end{figure}

\section{Curvature within the hyperbolic family}
\label{app:phase1}

The body \cref{fig:phase1} shows the dimension scaling for both
geometries; here we add detail on curvature. Sweep configuration:
geometry $\times$ $d \in \{2,4,8,16,32,64\}$ $\times$
$c \in \{0.1, 0.3, 1, 3\}$ for hyperbolic, three seeds. Within the
hyperbolic family, curvature interpolates between the two extremes:
higher $c$ pulls prototypes closer to the boundary of the disk,
where the Poincar\'e metric becomes most curved; this exchanges
$2$--$3$\,pp of top-1 for $4$--$5$\,pp of sibling recall. At $d{=}8$,
the optimal $c$ for top-1 is $0.3$ ($59.8\%$) and the optimal $c$
for sibling recall is $3.0$ ($0.230$). This is consistent with how
hyperbolic capacity is distributed: most volume lies near the
boundary, where the leaves of a tree should live.

\section{Regularizer-strength sweep, full panel set}
\label{app:phase2-full}

\begin{figure}[h]
  \centering
  \includegraphics[width=0.85\linewidth]{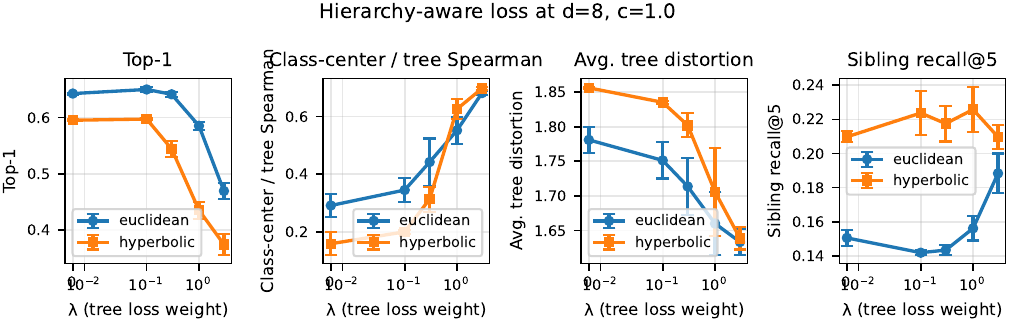}
  \caption{Full four-panel version of the regularizer-strength sweep
  (body \cref{fig:phase2} drops the third panel for legibility).
  Mean tree distortion decreases monotonically with $\lambda$ in
  both geometries, mirroring the rise in tree-Spearman.}
  \label{fig:phase2-full}
\end{figure}

\section{Training-tree ablation, full panel set}
\label{app:phase3-full}

The body \cref{fig:phase3} summarises the three-tree ablation; the
underlying observation is that training against a tree other than
the evaluation tree provides essentially no useful hierarchical
signal: tree-Spearman against the default tree drops to
$\approx 0.14$ for both non-default training trees, and top-1 climbs
back to the $\lambda{=}0$ baseline because the regularizer no longer
competes with the likelihood.

\begin{figure}[h]
  \centering
  \includegraphics[width=0.85\linewidth]{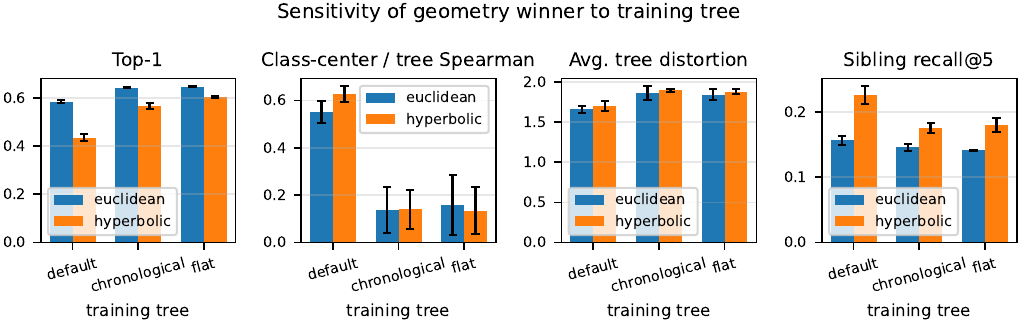}
  \caption{Full four-panel version of the training-tree ablation
  (body \cref{fig:phase3} drops the third panel for legibility).
  Tree distortion against the default tree is uniformly higher when
  training against a non-default tree.}
  \label{fig:phase3-full}
\end{figure}

\section{Empirical reference trees and DINOv2 cross-encoder}
\label{app:phase45}

\begin{figure}[h]
  \centering
  \includegraphics[width=0.85\linewidth]{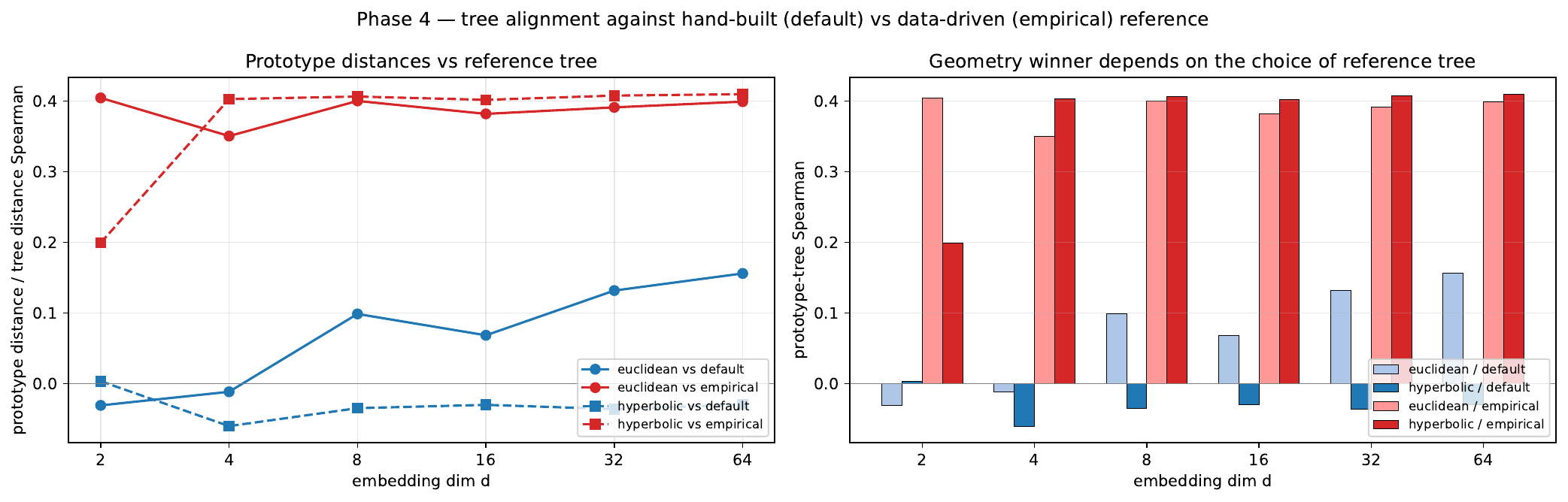}
  \caption{Prototype-distance / tree-distance Spearman against the
  hand-built default tree (blue) and a CLIP-empirical tree (red).
  Both geometries' prototypes align substantially better with the
  empirical tree than with the hand-built one. Caveat: the
  empirical tree is built in Euclidean space from the same CLIP
  features used to train the prototypes, so the alignment with
  prototype distances is partly expected by construction. The
  geometry comparison is the load-bearing claim, not the absolute
  level.}
  \label{fig:phase4}
\end{figure}

\begin{figure}[h]
  \centering
  \includegraphics[width=0.85\linewidth]{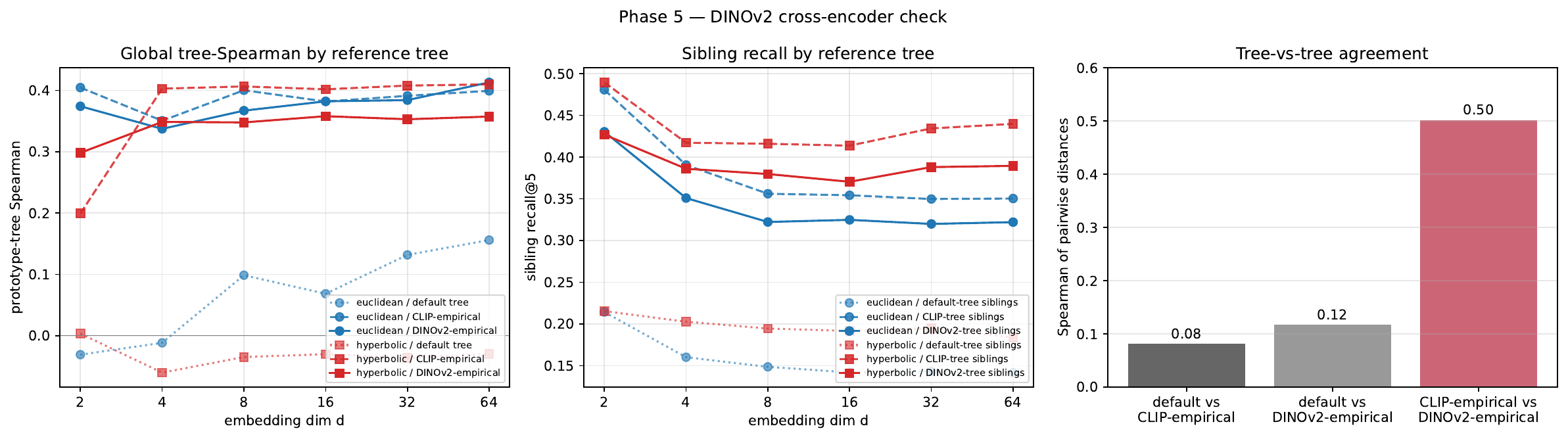}
  \caption{Substituting a DINOv2 ViT-B/14 encoder for CLIP in the
  empirical-tree construction. \textbf{Left:} prototype-tree
  Spearman against three references. The hyperbolic global-Spearman
  lead seen against the CLIP-empirical tree does not replicate
  against the DINOv2-empirical tree: Euclidean prototypes lead at
  every $d \geq 4$. \textbf{Centre:} sibling recall@5 against the
  three reference-tree sibling-set definitions. Hyperbolic leads on
  every reference at every $d \geq 4$. \textbf{Right:}
  pairwise-distance Spearman between reference trees. Hand-built
  correlates only weakly with either encoder-derived tree
  ($0.08$, $0.12$); the two encoder-derived trees correlate $0.50$
  with each other.}
  \label{fig:phase5}
\end{figure}

\paragraph{Tree-vs-tree.} Pairwise-distance Spearman between
hand-built default and CLIP-empirical is $0.08$; between default and
DINOv2-empirical it is $0.12$; between CLIP-empirical and
DINOv2-empirical it is $0.50$. Whatever the hand-built lineage tree
captures, the encoders see only a weak projection of it; whatever
the encoders capture, they capture it similarly. The hand-built
tree is the outlier.

\paragraph{Empirical sibling and cousin recall.} Sibling/cousin sets
used by recall@5 in the body figures are derived from
\texttt{STYLE\_HIERARCHY} (the default tree). To test reference-tree
sensitivity we rebuild sibling and cousin sets from each empirical
tree's binary linkage (siblings: the leaves in the other branch of
the leaf's first merge; cousins: the leaves in the uncle subtree at
the grandparent merge) and rerun the recall computation. The hyperbolic lead is preserved on every reference
tree (sibling recall@5 at $d{=}8$, default / CLIP-empirical /
DINOv2-empirical: Eu $0.149 / 0.356 / 0.322$, Hy $0.195 / 0.416 /
0.380$).

\paragraph{Global-Spearman direction is encoder-specific.} Against
the DINOv2-empirical tree, Euclidean prototypes lead at every
$d \geq 4$, with $\rho_{\textrm{Eu}} - \rho_{\textrm{Hy}}$ growing to
$+0.06$ at $d{=}64$ ($0.413$ vs.\ $0.357$). This is a negative result
for the strongest reading of the CLIP-empirical finding and
motivates the global-fidelity caveat in the body.

\section{Class-imbalance robustness}
\label{app:imbalance}

WikiArt is heavily skewed (Impressionism has $133\times$ more
training examples than Action Painting). The default sweep uses
unweighted cross-entropy; rerunning the $d{=}8$ winners of each
geometry with inverse-frequency class-weighted cross-entropy (three
seeds) shifts absolute numbers as expected (top-1 drops by roughly
$6$\,pp for both: Eu $64.3 \to 58.3\%$, Hy $59.8 \to 53.4\%$; while
balanced accuracy rises sharply: Eu $56.4 \to 65.6\%$, Hy
$44.5 \to 62.1\%$), but the geometry gap is preserved on every axis. Top-1 remains $5$\,pp Euclidean, sibling recall remains
$4$\,pp hyperbolic, and class-center / tree Spearman shifts by less
than a hundredth. The balanced-accuracy gap narrows from $11.9$ to
$3.5$\,pp, suggesting hyperbolic's main classification weakness in
the default setup was disproportionately on rare classes; it remains
in Euclidean's favour.

\section{Confusion structure}
\label{app:confusion}

\begin{figure}[h]
  \centering
  \includegraphics[width=0.85\linewidth]{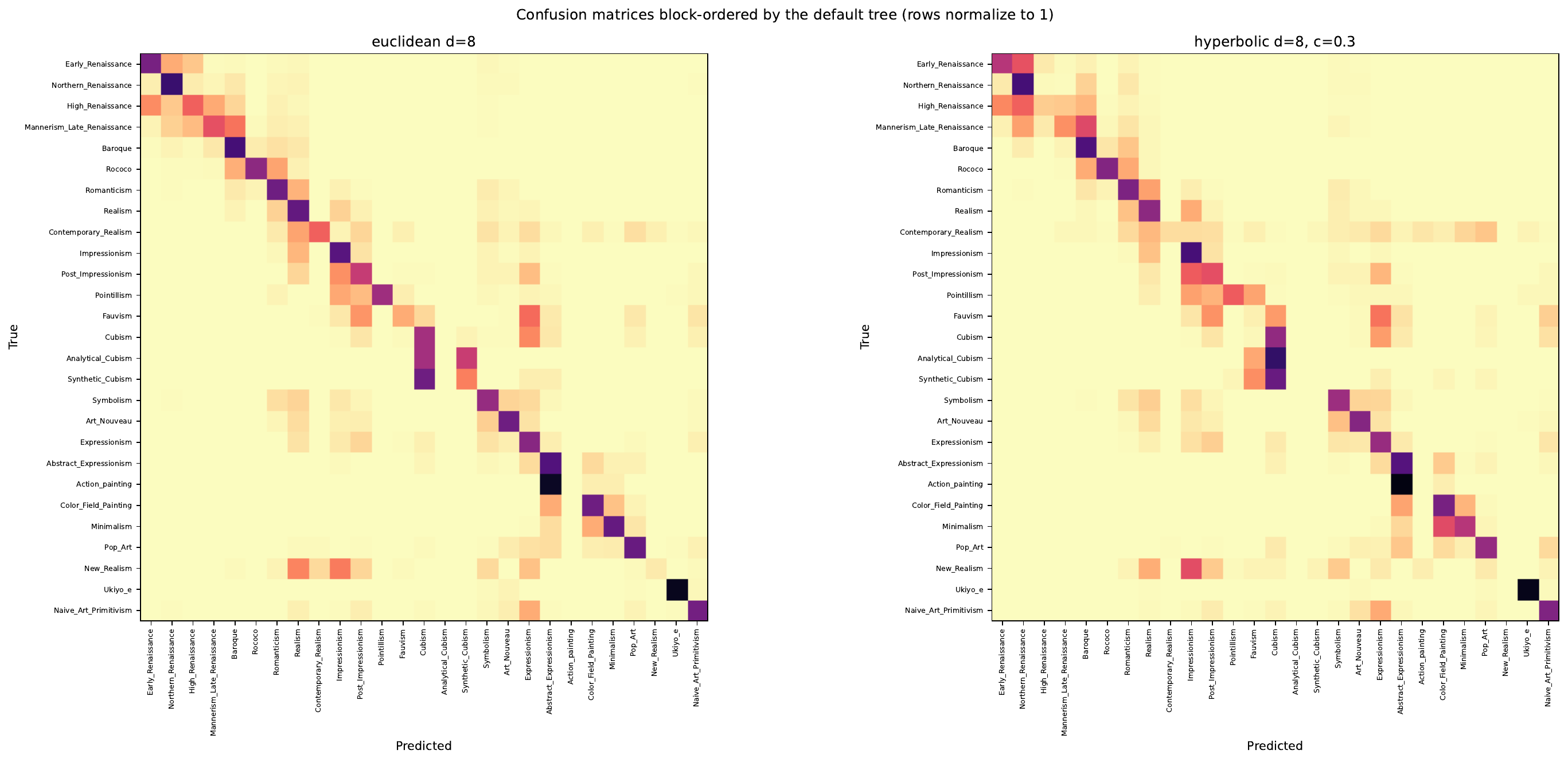}
  \caption{Row-normalized confusion matrices at $d{=}8$, with rows
  and columns permuted by a depth-first traversal of the default
  style tree so that hierarchically adjacent styles sit next to each
  other on both axes. Both models concentrate mistakes in
  near-diagonal blocks: errors predominantly fall on tree-adjacent
  styles. The hyperbolic block structure is visibly
  softer near the diagonal: more confusion mass spreads into
  immediate siblings, less mass jumps to distant styles. This is
  the same fact the sibling-recall numbers report, in a form a
  viewer can absorb at a glance.}
  \label{fig:confusion}
\end{figure}

\section{Reproducibility}
\label{app:repro}

The full sweep CSV records, for every config, the training
hyperparameters and every evaluation metric (150 rows). To reproduce:
\begin{verbatim}
python scripts/sweep.py --phase 1 --device mps  # ~14 min, 90 configs
python scripts/sweep.py --phase 2 --device mps  # ~10 min, 48 configs
python scripts/sweep.py --phase 3 --device mps  # ~3 min,  18 configs
python scripts/baselines.py                     # logreg, kNN baselines
python scripts/significance.py                  # paired-t, sign tests
python scripts/make_figures.py                  # regenerate figures
\end{verbatim}
Code, sweep configurations, and exact metric implementations are
available at
\url{https://github.com/pgrindehollevik-harvard/hyperbolic}.

\end{document}